\documentclass[runningheads]{llncs}
\usepackage[T1]{fontenc}
\usepackage{graphicx}
\usepackage{hyperref}
\usepackage{color}

\usepackage[table]{xcolor} 
\usepackage{multirow} 
\usepackage{xcolor} 
\usepackage{amsmath} 
\usepackage{amssymb} 
\usepackage[table]{xcolor} 
\definecolor{mygreen}{RGB}{0,150,0} 
\definecolor{myred}{RGB}{200,0,0} 
\usepackage[export]{adjustbox} 
\usepackage{cite} 
\usepackage{microtype} 
\usepackage{subcaption}
\begin{document}
%
\title{A Baseline Study and Benchmark for Few-Shot Open-Set Action Recognition with Feature Residual Discrimination}
%
\titlerunning{Few-Shot Open-Set Action Recognition}
%
\author{Stefano Berti\orcidID{0009-0002-0906-2251} \and
Giulia Pasquale\orcidID{0000-0002-7221-3553} \and
Lorenzo Natale\orcidID{0000-0002-8777-5233}}
\authorrunning{S. Berti et al.}
%
\institute{Humanoid Sensing and Perception, Istituto Italiano di Tecnologia, Genoa, Italy \\
\email{\{stefano.berti, giulia.pasquale, lorenzo.natale\}@iit.it}}
\maketitle              
\begin{abstract}
Few-Shot Action Recognition (FS-AR) has shown promising results but is often limited by a closed-set assumption that fails in real-world open-set scenarios. While Few-Shot Open-Set (FSOS) recognition is well-established for images, its extension to spatio-temporal video data remains underexplored. To address this, we propose an architectural extension based on a Feature-Residual Discriminator (FR-Disc), adapting previous work on skeletal data to the more complex video domain. Extensive experiments on five datasets demonstrate that while common open-set techniques provide only marginal gains, our FR-Disc significantly enhances unknown rejection capabilities without compromising closed-set accuracy, setting a new state-of-the-art for FSOS-AR. The project website, code, and benchmark are available at: \href{https://hsp-iit.github.io/fsosar/}{https://hsp-iit.github.io/fsosar/}.

\end{abstract}
\section{Introduction}
Deep learning has achieved remarkable success in action recognition by learning complex spatio-temporal representations from large-scale datasets \cite{pareek2021survey}.
However, the high cost of video annotation limits these models in real-world scenarios. Few-shot (FS) recognition addresses this by enabling generalization to novel classes with minimal labeled examples~\cite{perrett2021temporal,thatipelli2022spatio,xing2023boosting,wu2022motion,wang2023molo,wang2024hyrsm++}. 
Recent advances in FS-AR have further leveraged multi-modal settings—such as RGB, optical flow, or text—to improve robustness through complementary information~\cite{wang2021semantic,wanyan2023active,wang2024clip,tang2024semantic,wang2024cross}.

\begin{figure}[t]
  \centering
    \includegraphics[width=0.8\linewidth]{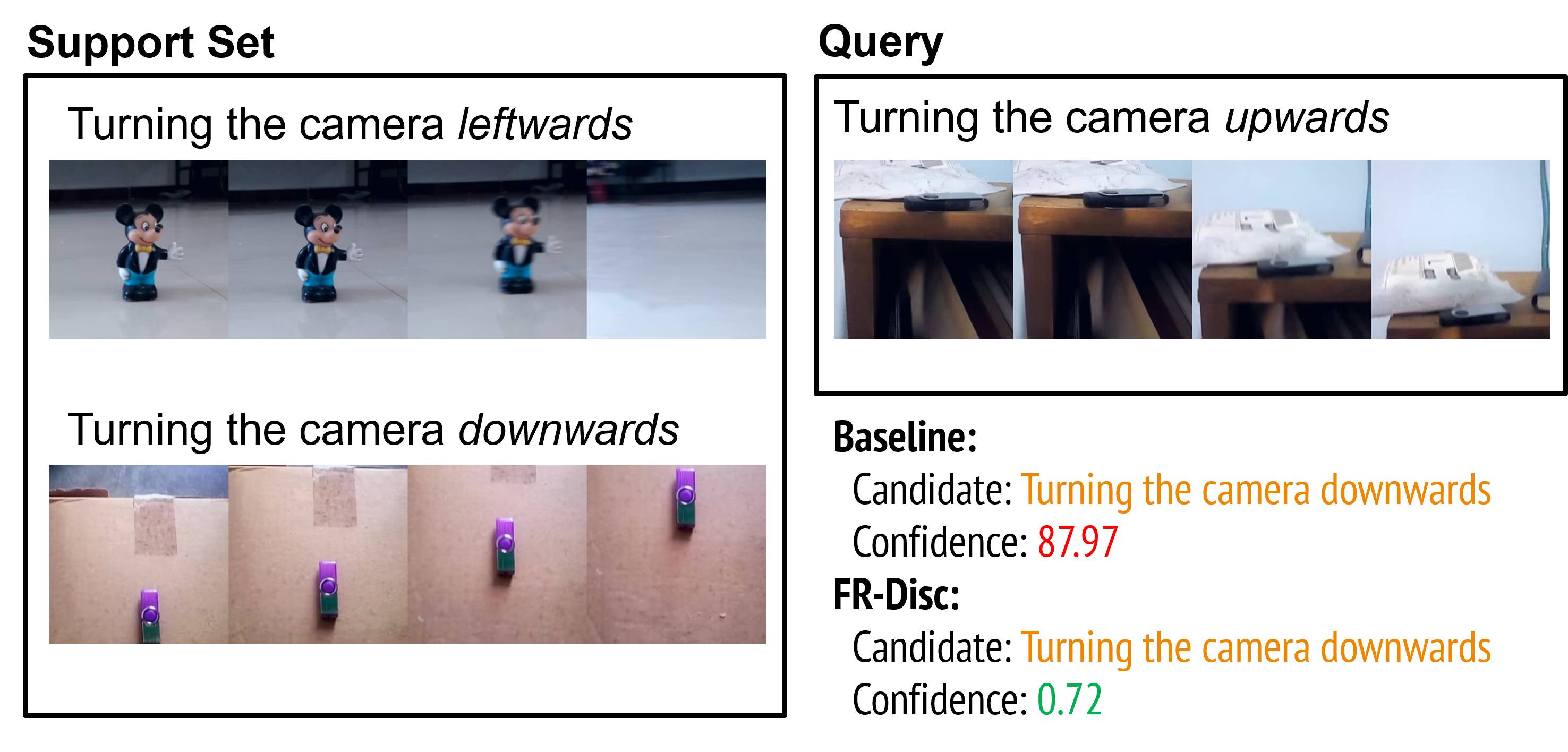}

   \caption{\textbf{Comparison between the results of the closed-set Softmax Baseline SAFSAR~\cite{tang2024semantic} and its proposed open-set extension FR-Disc on an unknown query}. The Support Set contains $K=2$ classes and $N=1$ videos. The true class of the query is not contained in the Support Set, so it should be rejected through low confidence (\textit{e.g.,} $< 50\%$). The Baseline incorrectly classifies it as known (false positive), while FR-Disc correctly rejects it (true negative). Additional qualitative examples are provided in the Supplementary Material. }
   \label{fig:main}
\end{figure}

A critical yet underexplored challenge in FS-AR is the open-set scenario, where models must reject unknown test samples to avoid false positives. While few-shot open-set (FSOS) recognition is established for images~\cite{liu2020few,jeong2021few,huang2022task,song2022few,deng2022learning,su2024toward,wang2023glocal,sun2024overall}, existing methods neglect the temporal dynamics of video. 
To our knowledge, the only existing approach for FSOS action recognition is our previous work~\cite{berti2022one}, which was limited to skeletal data.

Traditional open-set techniques rely either on scoring mechanisms—such as Maximum Softmax Score (MSS) or Maximum Logit Score (MLS)~\cite{dhamija2018reducing}—or auxiliary classification modules~\cite{ochal2024eol,wang2023glocal}. Studies in image-based OS recognition~\cite{roady2020open,vaze2021open,bisgin2024large} show that MLS often outperforms MSS and more complex methods because the Softmax operation discards logit magnitude, which is vital for uncertainty estimation~\cite{vaze2021open}. We investigate these insights in the video-centric FSOS domain by establishing a comprehensive benchmark suite across five datasets. We adapt two distinct FS-AR models, STRM~\cite{thatipelli2022spatio} and SAFSAR~\cite{tang2024semantic}, with these techniques, to provide a baseline for performance-efficiency trade-off.

Our main contributions are as follows: \begin{itemize} \item We adapt five state-of-the-art FS-AR video datasets to open-set conditions and establish the first comprehensive set of baselines for FSOS-AR. \item We demonstrate that FS-AR models can be equipped with OS recognition techniques without degrading closed-set accuracy, extending image-based OS findings to the video domain. \item We introduce a specialized architecture by evolving the Feature-Residual Discriminator (FR-Disc)~\cite{berti2022one} for high-dimensional video data. We show that modeling discrepancies between query features and class prototypes is superior to standard logit-based metrics (MLS/MSS) for capturing complex temporal dynamics. \end{itemize}

Overall, this work provides a foundational study of FSOS-AR, offering a robust evaluation of varied approaches on public benchmarks.

\section{Related work}
\subsection{Few-Shot Action Recognition} FS-AR advances focus on modeling spatio-temporal dynamics to generalize across novel classes.
Temporal-relational methods like TRX~\cite{perrett2021temporal} and STRM~\cite{thatipelli2022spatio} align query-support patterns via cross-transformers, while motion-aware approaches such as MTFAN~\cite{wu2022motion} and MoLo~\cite{wang2023molo} leverage motion-modulated fragments or contrasting contexts. Other frameworks integrate graph-guided matching (GgHm \cite{xing2023boosting}) or temporal set alignment (HyRSM++~\cite{wang2024hyrsm++}). 
Multi-modal trends further enhance robustness by using semantic priors~\cite{wang2021semantic}, commonsense knowledge prompting~\cite{shi2022knowledge,shi2024commonsense}, multi-modal feature exploration~\cite{wanyan2023active}, or CLIP-guided modulation~\cite{wang2024clip,wang2024cross}. Specifically, SAFSAR~\cite{tang2024semantic} demonstrates state-of-the-art performance in 1-shot settings. Despite their success, these models assume a closed-set environment, limiting real-world utility.

\subsection{Open-Set Recognition} \noindent{\textbf{Images.}} Early works introduce training-based losses like Entropic Open-Set (EOS) and Objectosphere to regularize feature magnitudes~\cite{dhamija2018reducing}. While foundational baselines relied on Maximum Softmax Scores (MSS) or ``garbage'' classes, recent studies~\cite{vaze2021open,bisgin2024large} show that Maximum Logit Scores (MLS) often achieve state-of-the-art results by preserving logit magnitude, outperforming generative~\cite{neal2018open,chen2021adversarial} or post-processing methods like OpenMax~\cite{ge2017generative}. Additionally, while input perturbations (e.g., ODIN) can improve detection, MLS remains a more practical baseline due to its computational efficiency~\cite{roady2020open}.

\noindent{\textbf{Videos.}} Video-centric OS recognition is less explored.
Recent works evaluate metrics like AUROC on datasets such as FineGym and Diving48 using Regularized-Discriminative OpenMax~\cite{yin2024rd}, or employ Evidential Deep Learning~\cite{bao2021evidential} and Information Bottleneck theory~\cite{cen2023enlarging} on UCF101 and HMDB51.
Skeleton-based OS-AR has also been recently benchmarked on NTURGBD~\cite{peng2024navigating}.

\subsection{Few-Shot Open-Set Recognition} \noindent{\textbf{Images.}} FSOS image recognition balances limited labels with unknown rejection through meta-learning (PEELER~\cite{liu2020few}), transformation consistency (SNATCHER \cite{jeong2021few}), task-adaptive sampling (SEMAN-G~\cite{huang2022task}), and background cues (ProCAM \cite{song2022few}).
Subsequent refinements include relative displacement~\cite{deng2022learning}, energy-based frameworks~\cite{wang2023glocal}, and vision-language integration~\cite{miller2024open,chen2024dual,sun2024overall,ochal2024eol}. These, however, do not encode the temporal information necessary for video.

\noindent{\textbf{Videos.}} To our knowledge, only our previous work~\cite{berti2022one} addresses FSOS-AR, using skeletal data and a feature-residual rejection module. This present work evolves that concept to high-dimensional video data, establishing its effectiveness on our proposed benchmark suite.

\section{Problem Definition and Metrics}
\subsection{Task Definition}We define Few-Shot Action Recognition (FS-AR) as a classification task where each class represents a different action. Let $D = \{(x_i,y_i) : y_i \in Y\}$ be a video dataset with label set $Y$, where $x_i$ and $y_i$ represent a video and its class respectively. The label set is split into disjoint training $Y^{\text{train}}$ and test $Y^{\text{test}}$ sets.

\noindent{\textbf{Few-Shot Action Recognition.}} Following episodic training~\cite{cao2020few,perrett2021temporal}, we sample $N^{\text{train}}$ tasks to form the task set $T^{\text{train}}_{\text{known}} = \{SS_i, Q_i^{\text{known}}\}_{i=1}^{N^{\text{train}}}$. Each support set, defined as \mbox{$SS_{i} = \{(x_{ij}, y_{ij}) : y_{ij} \in Y_i^{SS}\}_{j=1}^{NK}$}, contains $NK$ examples, that are $N$ videos for each of the $K$ classes sampled from $Y^{\text{train}}$. The query $Q_i^{\text{known}}$ belongs to the same label set $Y^{SS}_i$. The objective is to train a classifier $f$ such that:\begin{equation}\label{eq:fs}P_{\text{FS}} \triangleq f(SS_i, x_i) = \hat{y}_i : \hat{y}i = y_i .\end{equation}At test time, the protocol uses tasks $T^{\text{test}}_{\text{known}}$ sampled from $Y^{\text{test}}$.

\noindent{\textbf{Few-Shot Open-Set Action Recognition.}} We extend this to FSOS-AR by introducing an ``unknown'' task set $T_{\text{unknown}}^{\text{train}} = \{SS_i, Q_i^{\text{unknown}}\}_{i=1}^{N^T}$, where the query $Q_i^\text{unknown}$ belongs to classes $Y^{\text{unknown}}_i = Y^{\text{train}}-Y^{SS}_i$ (or $Y^{\text{test}}-Y^{SS}_i$ during testing). The full task set is $T = T_{\text{known}} \cup T_{\text{unknown}}$. We first define a binary classification problem $P_{\text{OS}}$ to distinguish known from unknown samples:\begin{equation}\label{eq:desired-f3}P_{\text{OS}} \triangleq f(SS_i, x_i) = \hat{a}_i : \hat{a}i = a_i ,\end{equation}where $a_i=1$ (accept) for known and $a_i=0$ (reject) for unknown queries. Ultimately, \textbf{FSOS-AR} requires predicting both the acceptance state and, if known, the specific class:
\begin{equation}
    \label{eq:fsos}
P_{\text{FSOS}} \triangleq f(SS_i, x_i) =
    \begin{cases}
        \hat{a}_i : \hat{a}_i = a_i \\
        \hat{y}_i : \hat{y}_i = y_i & \text{if } a_i=1
    \end{cases}
\end{equation}

\subsection{Performance Metrics}
We evaluate performance using three categories of metrics:

\noindent{\textbf{Closed-Set Metrics.}} Few-shot accuracy (\textbf{FS ACC}) evaluates the model's ability to discriminate among $K$ known classes ($P_{FS}$, Eq.~\ref{eq:fs}), ignoring open-set detection.

\noindent{\textbf{Open-Set Metrics.}} To evaluate the $P_{OS}$ binary task (Eq.~\ref{eq:desired-f3}), we report open-set accuracy (\textbf{OS ACC}), \textbf{AUROC}, and \textbf{AUPR}. These are computed over the balanced set $T_\text{known}^\text{test} \cup T_\text{unknown}^\text{test}$.

\noindent{\textbf{Joint Metrics.}} The Open-Set Classification Rate (\textbf{OSCR})~\cite{dhamija2018reducing} measures the trade-off between open-set detection and closed-set accuracy (Eq.~\ref{eq:fsos}). It computes the area under the Correct Classification Rate (CCR) vs. False Positive Rate (FPR) curve, where $\text{CCR} = \frac{1}{|T_\text{known}^\text{test}|} \sum_{(SS_i, Q_i) \in T_\text{known}^\text{test}} \mathbb{I}[\hat{y}_i = y_{i}] \mathbb{I}[ \hat{a}_i = 1 ]$.

\section{Methods}
\label{sec:methods}

\begin{figure*}[t]
  \centering
  \includegraphics[scale=0.15]{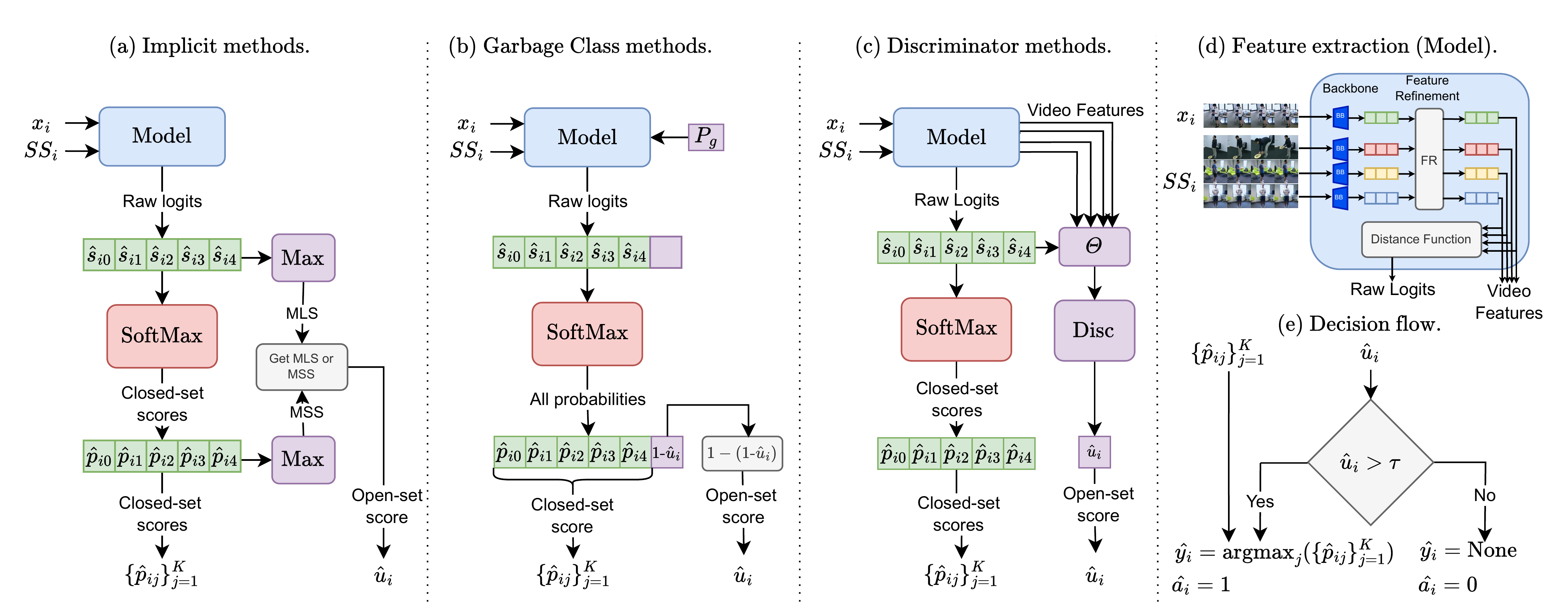}
  \caption{Overview of closed- and open-set prediction flows for (a) implicit (Softmax, EOS), (b) explicit (Garbage Class), and (c) Discriminator-based methods. For implicit methods, $\hat{u}_i$ is derived via MLS/MSS. In (b), a query is rejected if the garbage prototype $P_g$ yields the highest probability. In (c), $\hat{u}_i$ is computed by an MLP with Sigmoid activation using the $\varTheta$ function (Eq.~\ref{eq:getminnorm}). (d) and (e) illustrate the feature extraction and inference decision flow for SAFSAR and STRM models.}
  \label{fig:methods}
\end{figure*}

\subsection{Few-Shot Action Recognition Models}
We employ two representative closed-set FS-AR baselines: \textbf{STRM}~\cite{thatipelli2022spatio}, a 2D transformer-based model, and \textbf{SAFSAR}~\cite{tang2024semantic}, a 3D model. 
STRM enhances 2D features via patch-level and frame-level aggregation modules to capture local and temporal details. 
SAFSAR utilizes a 3D VideoMAE~\cite{tong2022videomae} backbone and incorporates text-based semantic priors to guide feature learning, offering superior performance at a higher computational cost. 
Both models use a standard Cross-Entropy (CE) loss on the known task set $T_\text{known}^\text{train}$:
\begin{equation}
    \mathcal{L}_{\text{CE}} = -\sum_{j=1}^{K} y_{ij} \log(\hat{p}_{ij})
\end{equation}
where $\hat{p}_{ij}$ is the predicted probability for class $j$. 
At inference, the class with the maximum probability is selected.

\subsection{Open-Set Techniques}
\label{sec:os}
We extend these baselines to output a known/unknown score $\hat{u}_i \in [0,1]$. We categorize these extensions into implicit and explicit techniques (Fig.~\ref{fig:methods}).

\subsubsection{Implicit OS Techniques}
These require no additional parameters, modifying only the scoring or loss function.
\begin{itemize}
    \item \textbf{Softmax Baseline}: Uses the Maximum Logit Score \textbf{(MLS)}, defined as \mbox{$\hat{u}_i = \text{Max}(\{\hat{s}_{ij}\}_{j=1}^{K})$} or the Maximum Softmax Score \textbf{(MSS)}, defined as  \mbox{$\hat{u}_i = \text{Max}(\text{Softmax}(\{\hat{s}_{ij}\}_{j=1}^{K}))=\text{Max}(\{\hat{p}_{ij}\}_{j=1}^{K}))$}  as the open-set score $\hat{u}_i$. A query is accepted if $\hat{u}_i > \tau$. 
    
    \item \textbf{Entropic Open-Set (EOS)}: Adds an entropy-boosting term to the loss to push predictions for unknown tasks $T_\text{unknown}^\text{train}$ toward a uniform distribution: $\mathcal{L}_{\text{EOS}} = -\frac{1}{K} \sum_{j=1}^{K} \log (\hat{p}_{ij})$. The total loss is $\mathcal{L} = \mathcal{L}_{CE} + \alpha_{EOS} \mathcal{L}_{EOS}$, where $\alpha_{EOS}$ balances the two terms.
\end{itemize}

\subsubsection{Explicit OS Techniques}
These methods introduce extra parameters to produce a $(K+1)$-way output, where the additional logit represents the ``unknown'' class.
\begin{itemize}
    \item \textbf{Garbage Class (GC)}: A random learnable ``Garbage Prototype'' $P_g$ is added to the support set. During training, unknown queries are assigned the label $K+1$. At inference, the query is rejected if $\hat{p}_{i,K+1}$ is the maximum probability; the open-set score is $\hat{u}_i = 1 - \hat{p}_{i,K+1}$.
    \item \textbf{Feature-Residual Discriminator (FR-Disc)}~\cite{berti2022one}: This approach introduces a lightweight auxiliary network (the Discriminator) fed with the ``residual'' (difference) between the query features and the features of the closest support class. The Discriminator is trained to distinguish between known and unknown queries. During training, positive samples are defined as known queries correctly classified by the closed-set model, while an equal number of negative (unknown) samples are used to maintain class balance. At inference, the Discriminator’s confidence score, $\hat{u}_i$, serves as the known/unknown indicator, which is compared against a threshold $\tau$ to accept or reject a query. The training loss for the Discriminator is defined as:
    \begin{align}
    \mathcal{L}_{Disc}(\{\hat{p}_{ij}\}^{K}_{j=1}, y_i) &=
    \mathbb{I}(\hat{y}_i = y_i)
    \cdot \text{BCE}(d_i, 1) \notag \\
    &+ \mathbb{I}(y_i \in Y^\text{unknown}_i) \cdot \text{BCE}(d_i, 0),
    \end{align}
    where $\mathbb{I}$ is the indicator function, BCE is the Binary Cross-Entropy loss, and $d_i$ represents the feature residual:
    \begin{equation}
        d_i = \varTheta(x_i, SS_i, \{\hat{s_i}\}_j) = \phi(x_i) - \phi(SS_{i,\text{argmax}_j(\{\hat{p}_{ij} \}_{j=1}^K)}),
    \label{eq:getminnorm}
    \end{equation}
    Here, $\phi(\cdot)$ denotes the feature extractor, $x_i$ is the query, and $\textit{argmax}_j$ returns the value $j$ corresponding to the maximum value of its argument. The final joint loss is:
    \begin{equation}
        \mathcal{L} = \mathcal{L}_{CE} + \alpha_{Disc} \mathcal{L}_{Disc} .
    \end{equation}

    where $\alpha_{Disc}$ is a weighting hyperparameter. This approach shares similarities with metric-based methods, such as Siamese Networks, which utilize feature differences to measure similarity and have been shown to enhance performance in few-shot classification~\cite{koch2015siamese}. We adapt this technique to the video domain by defining individual Discriminators for both the SAFSAR and STRM modules, more details are available in Supplementary Material.
\end{itemize}

\section{Experimental Setup}

\subsection{FSOS-AR Benchmark Suite}
To establish a robust FSOS-AR benchmark, we utilize train/test splits from existing FS-AR literature to generate known and unknown task sets ($T^\text{train}, T^\text{test}$). We employ five datasets covering diverse domains:
\begin{itemize}
    \item \textbf{HMDB51}~\cite{kuehne2011hmdb} \& \textbf{UCF101}~\cite{kay2017kinetics}: Standard benchmarks for human actions. We use the splits from~\cite{zhang2020few} (HMDB: 31/10 train/test classes; UCF: 70/21).
    \item \textbf{SSv2}~\cite{goyal2017something}: Focuses on fine-grained object interactions. We adopt the split from~\cite{cao2020few} (64/24 classes).
    \item \textbf{NTURGBD}~\cite{shahroudy2016ntu} \& \textbf{Diving48}~\cite{li2018resound}: For NTU, we select two-person interactions (82 classes remaining). For both datasets, we generate random 70/30\% train/test splits (NTU: 57/25 classes; Diving48: 32/15 classes).
\end{itemize}

\subsection{Implementation Details}
Following standard FS-AR protocols, we evaluate on 5-Way, 1-Shot and 5-Shot tasks, sampling 8 frames per video.
We use the original hyperparameters for STRM~\cite{thatipelli2022spatio}. 
For SAFSAR~\cite{tang2024semantic}, we re-implemented the model using VideoMAE weights pre-trained on Kinetics-400~\cite{tong2022videomae}.
For the explicit thresholding metrics (OS ACC), we fix $\tau=0.5$. 
Other metrics (AUROC, OSCR) are threshold-independent.
Additional implementation details are provided in the Supplementary Material.

\section{Results}

\begin{table*}[t]
\fontsize{9pt}{10pt}\selectfont
\centering
\caption{Combined results for SAFSAR in 5-way 1-Shot and 5-Shot settings. Green/red values indicate improvement/decline over the Softmax baseline. Values marked with (*) denote models trained for 1K iterations to prevent overfitting.}
\label{table:safsar}
    \begin{tabular}{l|cc|cc|cc|cc|cc}
    \hline\hline
    \multirow{2}{*}{\textbf{OS-Method}} & \multicolumn{2}{c}{\textbf{FS ACC}} & \multicolumn{2}{c}{\textbf{OS ACC}} & \multicolumn{2}{c}{\textbf{AUROC}} & \multicolumn{2}{c}{\textbf{AUPR}} & \multicolumn{2}{c}{\textbf{OSCR}} \\
    ~ & \textbf{1-s} & \textbf{5-s} & \textbf{1-s} & \textbf{5-s} & \textbf{1-s} & \textbf{5-s} & \textbf{1-s} & \textbf{5-s} & \textbf{1-s} & \textbf{5-s} \\ 
    \hline
    \multicolumn{11}{c}{\cellcolor{gray!10}\textit{Diving48}} \\ \hline
    Softmax & 63.49 & 74.12 & 64.16 & 65.36 & 68.48 & 71.49 & 69.71 & 72.07 & 59.86 & 66.15 \\
    EOS & \textcolor{mygreen}{64.43} & \textcolor{myred}{72.80} & \textcolor{myred}{62.91} & \textcolor{myred}{64.94} & \textcolor{mygreen}{68.64} & \textcolor{mygreen}{74.60} & 69.71 & \textcolor{mygreen}{75.06} & \textcolor{mygreen}{60.36} & \textcolor{mygreen}{67.43} \\
    GC & \textcolor{mygreen}{65.01} & \textcolor{mygreen}{76.32} & \textcolor{mygreen}{64.89} & \textcolor{mygreen}{69.36} & \textcolor{mygreen}{70.18} & \textcolor{mygreen}{75.82} & \textcolor{mygreen}{71.38} & \textcolor{mygreen}{76.01} & \textcolor{mygreen}{60.63} & \textcolor{mygreen}{69.13} \\
    \textbf{FR-Disc} & \textcolor{mygreen}{\textbf{68.83}} & \textcolor{mygreen}{\textbf{78.58}} & \textcolor{mygreen}{\textbf{66.22}} & \textcolor{mygreen}{\textbf{70.29}} & \textcolor{mygreen}{\textbf{71.28}} & \textcolor{mygreen}{\textbf{76.55}} & \textcolor{mygreen}{\textbf{72.28}} & \textcolor{mygreen}{\textbf{76.32}} & \textcolor{mygreen}{\textbf{63.46}} & \textcolor{mygreen}{\textbf{71.04}} \\ \hline
    
    \multicolumn{11}{c}{\cellcolor{gray!10}\textit{SSv2}} \\ \hline
    Softmax & 62.11 & 74.08 & 62.29 & 69.61 & 70.39 & 77.05 & 73.02 & 78.80 & 60.72 & 69.35 \\
    EOS & \textcolor{mygreen}{62.97} & \textcolor{myred}{73.84} & \textcolor{mygreen}{65.08} & \textcolor{mygreen}{69.90} & \textcolor{mygreen}{71.56} & \textcolor{mygreen}{79.60} & \textcolor{mygreen}{73.74} & \textcolor{mygreen}{81.18} & \textcolor{mygreen}{61.53} & \textcolor{mygreen}{70.38} \\
    GC & \textcolor{mygreen}{62.47} & \textcolor{mygreen}{76.24} & \textcolor{mygreen}{64.06} & \textcolor{mygreen}{70.89} & \textcolor{myred}{69.20} & \textcolor{mygreen}{77.62} & \textcolor{myred}{70.47} & \textcolor{myred}{76.92} & \textcolor{myred}{59.26} & \textcolor{mygreen}{70.20} \\
    \textbf{FR-Disc} & \textcolor{mygreen}{\textbf{63.37}} & \textcolor{mygreen}{\textbf{77.88}} & \textcolor{mygreen}{\textbf{66.56}} & \textcolor{mygreen}{\textbf{73.52}} & \textcolor{mygreen}{\textbf{72.18}} & \textcolor{mygreen}{\textbf{81.56}} & \textcolor{mygreen}{\textbf{74.98}} & \textcolor{mygreen}{\textbf{82.96}} & \textcolor{mygreen}{\textbf{62.12}} & \textcolor{mygreen}{\textbf{73.18}} \\ \hline

    \multicolumn{11}{c}{\cellcolor{gray!10}\textit{NTURGBD}} \\ \hline
    Softmax & 88.31 & 91.58 & 79.90 & 81.45 & 87.76 & 91.44 & 88.35 & 91.30 & 81.45 & 84.83 \\
    EOS & \textcolor{myred}{87.63} & \textcolor{mygreen}{91.86} & \textcolor{mygreen}{80.10} & \textcolor{mygreen}{82.18} & \textcolor{mygreen}{88.17} & \textcolor{mygreen}{91.89} & \textcolor{mygreen}{88.96} & \textcolor{mygreen}{92.19} & \textcolor{myred}{81.34} & \textcolor{mygreen}{85.11} \\
    GC & \textcolor{mygreen}{89.07} & \textcolor{mygreen}{92.40} & \textcolor{mygreen}{81.78} & \textcolor{mygreen}{81.51} & \textcolor{mygreen}{89.30} & \textcolor{myred}{89.37} & \textcolor{mygreen}{88.94} & \textcolor{myred}{88.63} & \textcolor{mygreen}{81.93} & \textcolor{myred}{84.03} \\
    \textbf{FR-Disc} & \textcolor{mygreen}{\textbf{89.97}} & \textcolor{mygreen}{\textbf{95.54}} & \textcolor{mygreen}{\textbf{82.95}} & \textcolor{mygreen}{\textbf{86.53}} & \textcolor{mygreen}{\textbf{89.78}} & \textcolor{mygreen}{\textbf{94.31}} & \textcolor{mygreen}{\textbf{89.82}} & \textcolor{mygreen}{\textbf{94.26}} & \textcolor{mygreen}{\textbf{83.12}} & \textcolor{mygreen}{\textbf{88.31}} \\ \hline

    \multicolumn{11}{c}{\cellcolor{gray!10}\textit{HMDB51}} \\ \hline
    Softmax* & 65.29 & 79.68 & 64.44 & 72.40 & 70.76 & 81.79 & 73.87 & 83.33 & 62.23 & 74.26\\
    EOS* & \textcolor{mygreen}{69.19} & \textcolor{mygreen}{80.18} & \textcolor{mygreen}{66.23} & \textcolor{mygreen}{72.48} & \textcolor{mygreen}{74.60} & \textcolor{myred}{81.45} & \textcolor{mygreen}{77.13} & \textcolor{mygreen}{83.50} & \textcolor{mygreen}{65.74} & \textcolor{mygreen}{74.61} \\
    GC* & \textcolor{myred}{62.85} & \textcolor{myred}{76.74} & \textcolor{myred}{60.19} & \textcolor{myred}{64.30} & \textcolor{myred}{68.99} & \textcolor{myred}{76.45} & \textcolor{myred}{71.87} & \textcolor{myred}{78.93} & \textcolor{myred}{60.49} & \textcolor{myred}{70.42} \\
    \textbf{FR-Disc} & \textcolor{mygreen}{\textbf{72.38}} & \textcolor{mygreen}{\textbf{85.17}} & \textcolor{mygreen}{\textbf{68.87}} & \textcolor{mygreen}{\textbf{76.99}} & \textcolor{mygreen}{\textbf{77.48}} & \textcolor{mygreen}{\textbf{87.94}} & \textcolor{mygreen}{\textbf{80.30}} & \textcolor{mygreen}{\textbf{89.75}} & \textcolor{mygreen}{\textbf{68.79}} & \textcolor{mygreen}{\textbf{80.15}} \\ \hline

    \multicolumn{11}{c}{\cellcolor{gray!10}\textit{UCF101}} \\ \hline
    Softmax* & 95.04 & 98.32 & 80.59 & 91.25 & 94.55 & 98.03 & 95.04 & 98.32 & 88.31 & 91.57 \\
    EOS* & \textcolor{myred}{94.84} & \textcolor{mygreen}{98.78} & \textcolor{mygreen}{81.30} & \textcolor{myred}{89.32} & \textcolor{mygreen}{95.18} & \textcolor{mygreen}{98.31} & \textcolor{mygreen}{95.73} & \textcolor{mygreen}{98.50} & \textcolor{mygreen}{88.47} & \textcolor{mygreen}{91.97} \\
    GC* & \textcolor{myred}{79.98} & \textcolor{myred}{86.33} & \textcolor{myred}{59.88} & \textcolor{myred}{57.49} & \textcolor{myred}{75.15} & \textcolor{myred}{81.91} & \textcolor{myred}{77.29} & \textcolor{myred}{82.99} & \textcolor{myred}{70.85} & \textcolor{myred}{77.43} \\
    \textbf{FR-Disc} & \textcolor{mygreen}{\textbf{95.72}} & \textcolor{mygreen}{\textbf{99.28}} & \textcolor{mygreen}{\textbf{86.82}} & \textcolor{mygreen}{\textbf{91.52}} & \textcolor{mygreen}{\textbf{95.19}} & \textcolor{mygreen}{\textbf{98.89}} & \textcolor{mygreen}{\textbf{96.23}} & \textcolor{mygreen}{\textbf{99.08}} & \textcolor{mygreen}{\textbf{89.01}} & \textcolor{mygreen}{\textbf{92.49}} \\ 
    \hline\hline
    \end{tabular}
\end{table*}

Table~\ref{table:safsar} and Table~\ref{table:strm} report the results for SAFSAR and STRM, respectively, across all five datasets and the four open-set techniques. In Section~\ref{section:softmax} we discuss the result of the Softmax baseline, while in Section~\ref{section:osmethods} we analyze the performance of the different Open-Set techniques.

In our experiments, we observed that for SAFSAR, the Softmax Baseline (as well as the EOS and Garbage Class methods) exhibits a performance decline on the HMDB51 and UCF101 datasets if training is extended. This behavior is suggestive of overfitting; while training metrics saturate almost immediately, validation results show a gradual deterioration.
This trend may be attributed to the relatively low class cardinality and predominantly spatial nature of HMDB51 and UCF101, combined with the strong action-representation capabilities of the pretrained VideoMAE backbone. In these cases, the model rapidly learns to recognize the training actions and may over-specialize its weights for episodic tasks involving \textit{only} training classes. To ensure a fair evaluation and maintain generalization, we limited the training to 1K iterations for SAFSAR on HMDB51 and UCF101 for the Softmax, EOS, and GC techniques in Table~\ref{table:safsar}.
In contrast, the FR-Disc technique did not require early stopping, as it exhibited no signs of overfitting in our experiments.

We also do not report performance for STRM on the 1-Shot task in Table~\ref{table:strm} because in this scenario STRM suffers of feature bias~\cite{qin2024tfrs}.

\begin{table}[t]
\fontsize{9pt}{10pt}\selectfont 
\setlength{\tabcolsep}{3.5pt} 
\centering
\caption{Results for STRM in 5-way 5-Shot settings. Green/red values indicate improvement/decline over the Softmax baseline.}
\label{table:strm}
    \begin{tabular}{lccccc}
    \hline\hline
    \textbf{OS-Method} & \textbf{FS ACC} & \textbf{OS ACC} & \textbf{AUROC} & \textbf{AUPR} & \textbf{OSCR} \\ 
    \hline
    \multicolumn{6}{c}{\cellcolor{gray!10}\textit{Diving48}} \\ \hline
    Softmax & 74.74 & 52.91 & 58.92 & 63.25 & 63.61 \\ 
    EOS & \textcolor{myred}{73.58} & \textcolor{mygreen}{60.51} & \textcolor{mygreen}{67.01} & \textcolor{mygreen}{69.02} & \textcolor{mygreen}{65.63} \\ 
    GC & \textcolor{myred}{73.22} & \textcolor{mygreen}{53.76} & \textcolor{mygreen}{71.71} & \textcolor{mygreen}{72.56} & \textcolor{mygreen}{66.09} \\
    \textbf{FR-Disc} & \textcolor{mygreen}{\textbf{77.74}} & \textcolor{mygreen}{\textbf{64.92}} & \textcolor{mygreen}{\textbf{75.73}} & \textcolor{mygreen}{\textbf{75.72}} & \textcolor{mygreen}{\textbf{70.00}} \\ \hline

    \multicolumn{6}{c}{\cellcolor{gray!10}\textit{SSv2}} \\ \hline
    Softmax & 65.25 & 53.98 & 61.94 & 66.33 & 60.77 \\ 
    EOS & \textcolor{myred}{63.61} & \textcolor{mygreen}{59.91} & \textcolor{mygreen}{64.29} & \textcolor{mygreen}{69.09} & \textcolor{myred}{60.65} \\ 
    GC & \textcolor{myred}{43.05} & \textcolor{myred}{50.68} & \textcolor{myred}{58.15} & \textcolor{myred}{62.45} & \textcolor{myred}{47.06} \\
    \textbf{FR-Disc} & \textcolor{mygreen}{\textbf{65.51}} & \textcolor{mygreen}{\textbf{58.99}} & \textcolor{mygreen}{\textbf{70.82}} & \textcolor{mygreen}{\textbf{73.59}} & \textcolor{mygreen}{\textbf{62.52}} \\ \hline

    \multicolumn{6}{c}{\cellcolor{gray!10}\textit{NTURGBD}} \\ \hline
    Softmax & 95.14 & 50.85 & 74.69 & 78.26 & 80.28 \\ 
    EOS & \textcolor{myred}{92.08} & \textcolor{mygreen}{75.80} & \textcolor{mygreen}{85.74} & \textcolor{mygreen}{86.32} & \textcolor{mygreen}{83.30} \\ 
    GC & \textcolor{myred}{85.71} & \textcolor{mygreen}{59.29} & \textcolor{mygreen}{80.27} & \textcolor{mygreen}{80.02} & \textcolor{myred}{76.84} \\
    \textbf{FR-Disc} & \textcolor{myred}{\textbf{93.28}} & \textcolor{mygreen}{\textbf{80.76}} & \textcolor{mygreen}{\textbf{92.27}} & \textcolor{mygreen}{\textbf{91.95}} & \textcolor{mygreen}{\textbf{86.11}} \\ \hline

    \multicolumn{6}{c}{\cellcolor{gray!10}\textit{HMDB51}} \\ \hline
    Softmax & 75.92 & 52.18 & 70.49 & 65.29 & 67.34 \\ 
    EOS & \textcolor{myred}{75.18} & \textcolor{mygreen}{64.99} & \textcolor{mygreen}{75.18} & \textcolor{mygreen}{70.58} & \textcolor{mygreen}{69.33} \\ 
    GC & \textcolor{myred}{71.14} & \textcolor{myred}{51.10} & \textcolor{mygreen}{78.85} & \textcolor{mygreen}{73.85} & \textcolor{myred}{65.96} \\
    \textbf{FR-Disc} & \textcolor{myred}{\textbf{75.50}} & \textcolor{mygreen}{\textbf{56.87}} & \textcolor{mygreen}{\textbf{77.59}} & \textcolor{mygreen}{\textbf{75.20}} & \textcolor{mygreen}{\textbf{69.11}} \\ \hline

    \multicolumn{6}{c}{\cellcolor{gray!10}\textit{UCF101}} \\ \hline
    Softmax & 95.48 & 51.20 & 86.79 & 89.39 & 87.40 \\ 
    EOS & \textcolor{myred}{95.05} & \textcolor{mygreen}{75.71} & \textcolor{mygreen}{90.35} & \textcolor{mygreen}{92.42} & \textcolor{myred}{86.87} \\ 
    GC & \textcolor{mygreen}{95.58} & \textcolor{mygreen}{63.21} & \textcolor{mygreen}{93.35} & \textcolor{mygreen}{95.34} & \textcolor{myred}{83.71} \\
    \textbf{FR-Disc} & \textcolor{mygreen}{\textbf{95.76}} & \textcolor{mygreen}{\textbf{63.47}} & \textcolor{mygreen}{\textbf{93.60}} & \textcolor{mygreen}{\textbf{94.56}} & \textcolor{mygreen}{\textbf{88.37}} \\ 
    \hline\hline
    \end{tabular}
\end{table}

\subsection{Evaluating the Softmax Baseline}
\label{section:softmax}
\begin{figure*}[t]
    \centering
    \begin{subfigure}[b]{0.48\textwidth}
        \centering
        \includegraphics[width=\textwidth, valign=m]{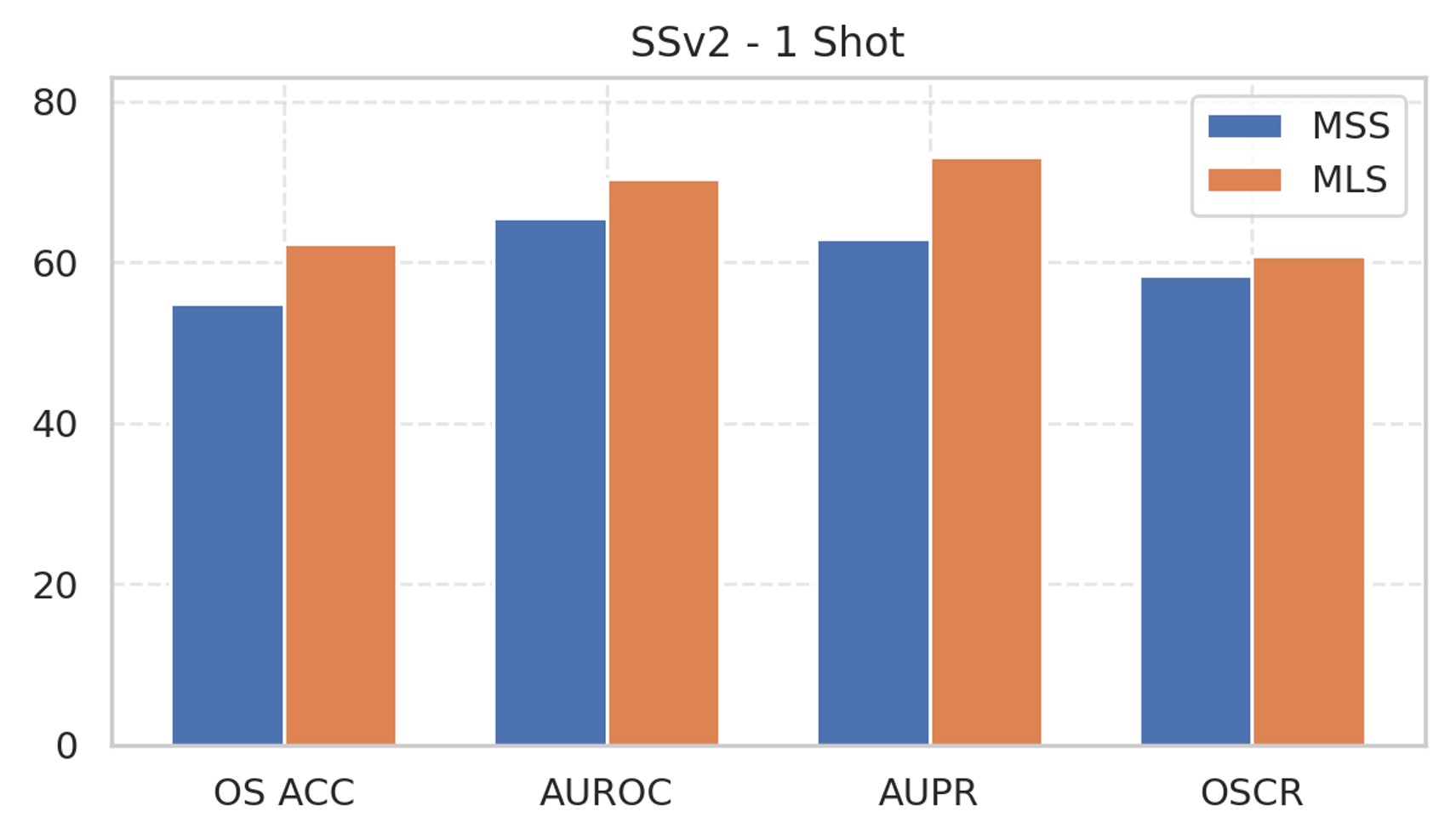}
        \caption{MSS vs. MLS comparison.}
        \label{fig:mls-vs-mss}
    \end{subfigure}
    \hfill
    \begin{subfigure}[b]{0.48\textwidth}
        \centering
        \includegraphics[width=\textwidth, valign=m]{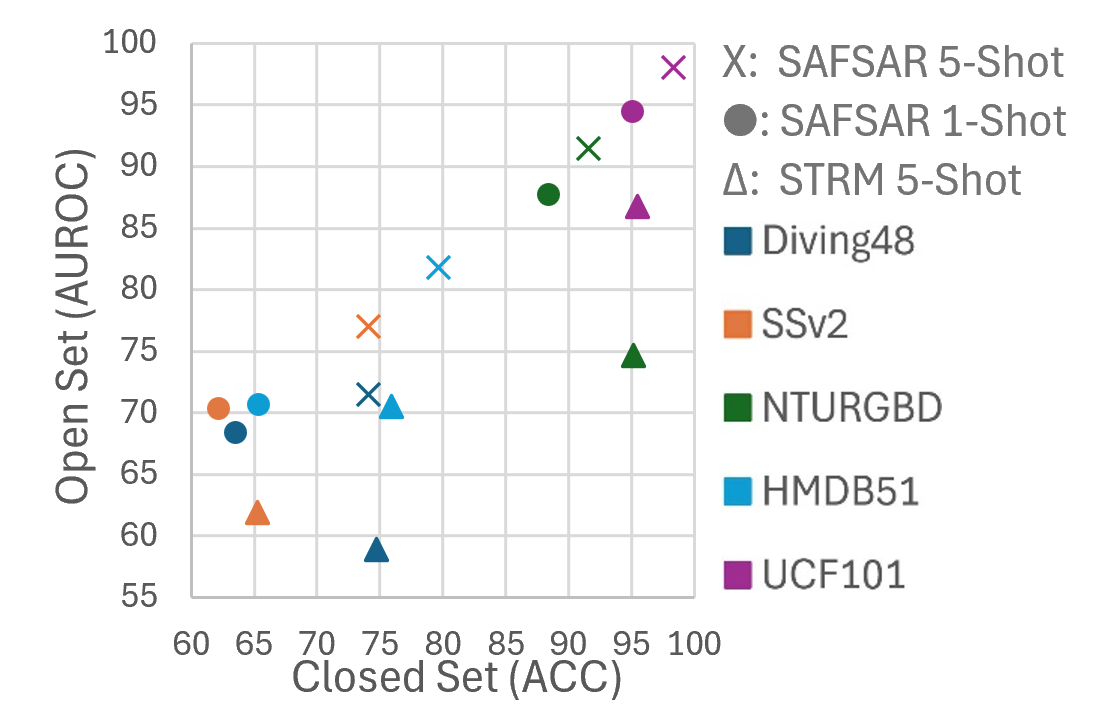}
        \caption{Metric correlation.}
        \label{fig:corr}
    \end{subfigure}
    \caption{Analysis of open-set scoring and performance correlation. (a) Comparison between MSS and MLS for SAFSAR's Softmax Baseline on SSv2 1-Shot. (b) Global correlation between closed-set and open-set metrics across all five datasets for both SAFSAR and STRM.}
\end{figure*}

\textbf{Logits normalization worsens open-set recognition.}
Previous works \cite{vaze2021open,bisgin2024large} have shown that MLS outperforms MSS in rejecting unknown queries in the image domain, as the Softmax normalization hides magnitude information.
We first investigated this claim in the spatio-temporal setting as well. 
In Figure~\ref{fig:mls-vs-mss} we compare the open-set performance of MLS and MSS for SAFSAR's Softmax Baseline (1-Shot) on the SSv2 dataset.
The result supports the claim in this domain too: MLS consistently outperforms MSS, underscoring the importance of preserving the logits magnitude for representing prediction confidence.
We did not conduct this comparison for STRM because MLS requires bounded logits, such as the cosine similarities produced by SAFSAR. In contrast, STRM produces unbounded logits (negative norms of feature differences). While normalization based on training set ranges is possible, it would necessitate additional threshold tuning; to ensure a fair comparison and avoid extra hyperparameter dependencies, we opt for the most compatible scoring method for each architecture.
Consequently, we employ MLS for SAFSAR and MSS for STRM with the Softmax baseline and EOS techniques.

\textbf{Closed-set models can be strong competitors in open-set tasks.} In Table~\ref{table:safsar} and Table~\ref{table:strm} we report the performance of respectively the SAFSAR and the STRM models combined with the considered OS techniques across datasets. 
We first observe that the Softmax Baseline achieves competitive open-set metrics. This is somewhat unexpected, as unknown queries can still resemble support examples and lead to confident false positives.
However, this also aligns with the claim made by~\cite{vaze2021open}, which suggests that a classifier’s ability to make ``none-of-the-above'' decisions (i.e., to detect unknown queries) is strongly correlated with its accuracy on the closed-set classes, likely due to their ability to learn discriminative, well-separated features.
This is particularly evident on the easiest dataset UCF101 where the overall performance is very high, especially for SAFSAR, and the gains offered by OS techniques over the closed-set baseline are smaller.

\textbf{Are closed- and open-set performance correlated?}
To investigate this finding in this spatio-temporal task, we conducted an experiment similar to~\cite{vaze2021open}. In ~\ref{fig:corr} we plot the closed- and open-set metrics in a scatter plot, 
for SAFSAR's and STRM's Softmax Baseline on all datasets.
We first notice, for each model, a linear correlation between these two metrics across datasets, with Pearson coefficients of 0.84 for STRM, and 0.99 and 0.97 for 1- and 5-Shot SAFSAR respectively. 
Moreover, SAFSAR's equal or better closed-set performance is coupled with a consistently better open-set performance with respect to STRM: this can be noticed when comparing SAFSAR 5-Shot (cross) vs. STRM 5-Shot (triangle) on a same dataset (same color) and noticing that the cross appears to the right or aligned to the triangle for each dataset except NTURGBD, and consistently above the triangle for each dataset.
This indicates that even without dedicated open-set mechanisms, a better-trained closed-set classifier can effectively generalize better to distinguish known and unknown classes.
We expect this correlation for the prototype-based methods considered here, as both closed- and open-set performance are rooted in the same discriminative feature learning process. Specifically, because the model relies on the same learned representations for both classification and outlier detection, improvements in feature clustering for known classes naturally enhance the ability to isolate unknown samples.


\subsection{Evaluating Open-set Techniques}
\label{section:osmethods}
\noindent \textbf{EOS: Benefits of exposing a closed-set model to unknowns.}
This is a relatively easy technique (requiring only tuning the loss weight $\sigma_{EOS}$) to expose the model to unknown queries, where it forces the output probabilities $\{\hat{p}_{ij}\}_{j=1}^K$ to have equal values $1 / K$ over all classes.
When EOS is applied to SAFAR in Table~\ref{table:safsar}, it occasionally causes minor drops in FS ACC ($-1.32\%$ on Diving48 5-Shot), but consistently improves OSCR metric across almost all datasets for both 1- and 5-Shot tasks.
EOS benefit is larger on the less powerful STRM in Table~\ref{table:strm}, providing improvements of ${\sim}10$-$25\%$ on the OS ACC on NTURGBD, UCF101 and HMDB51, and an average ${\sim}5-10\%$ gain on the AUROC and AUPR metrics. As for SAFSAR, the gain on the open-set metrics comes with a minor cost on the FS ACC, with the highest reduction of ${\sim}3\%$ on NTURGBD. 
This result shows that good open-set rejection improvements can be obtained with this simple technique on more or less powerful prototype-based models as SAFSAR and STRM, with a minor cost on the closed-set accuracy.

\begin{figure*}[t]
    \centering
    \begin{subfigure}[b]{0.48\textwidth}
        \centering
        \includegraphics[width=0.49\textwidth]{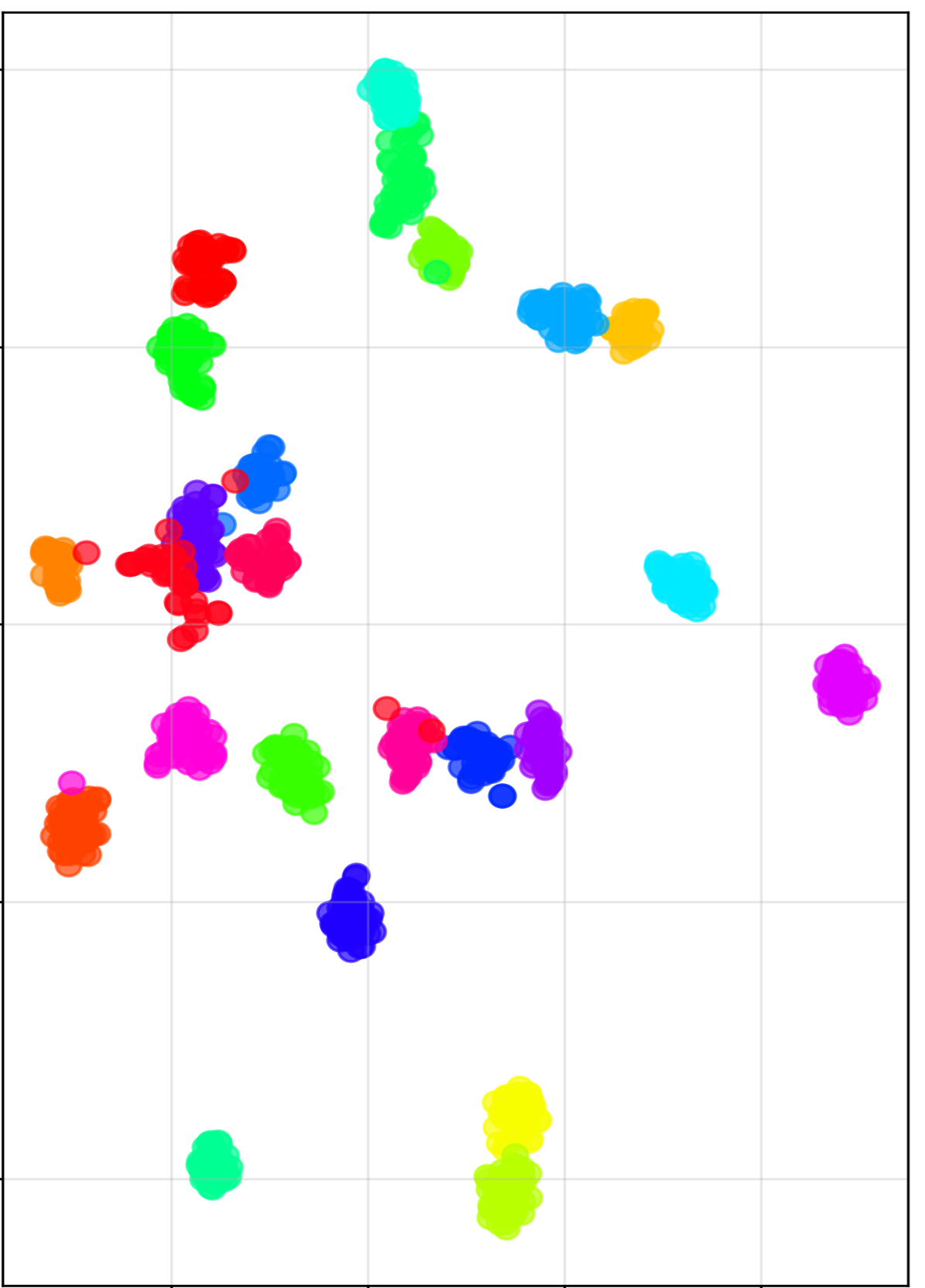}
        \hfill
        \includegraphics[width=0.49\textwidth]{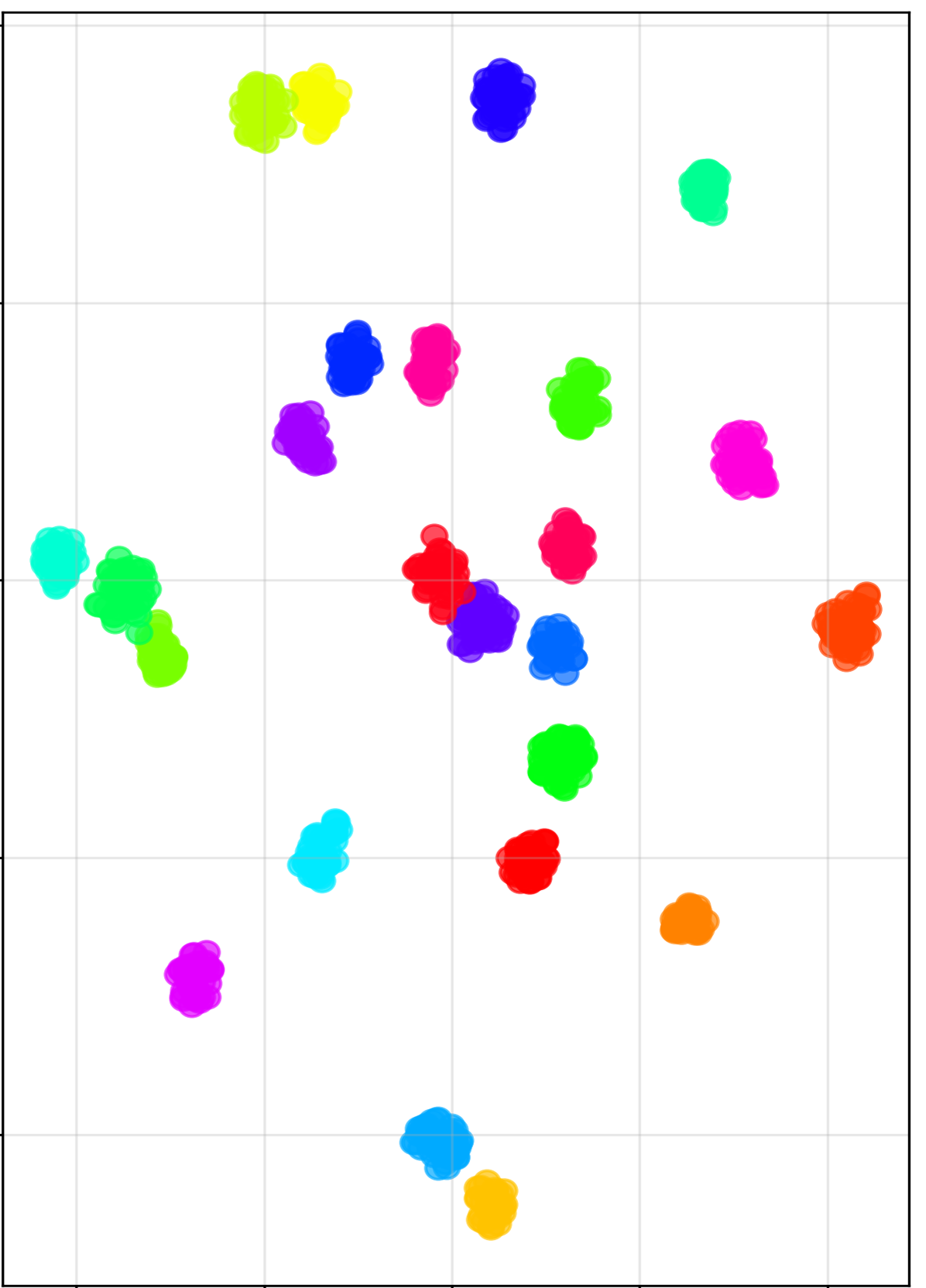}
        \caption{Feature t-SNE visualizations}
        \label{fig:tsne_group}
    \end{subfigure}
    \hfill
    \vrule 
    \hfill
    \begin{subfigure}[b]{0.48\textwidth}
        \centering
        \includegraphics[width=0.85\textwidth, trim={0pt 0 0 0}, clip]{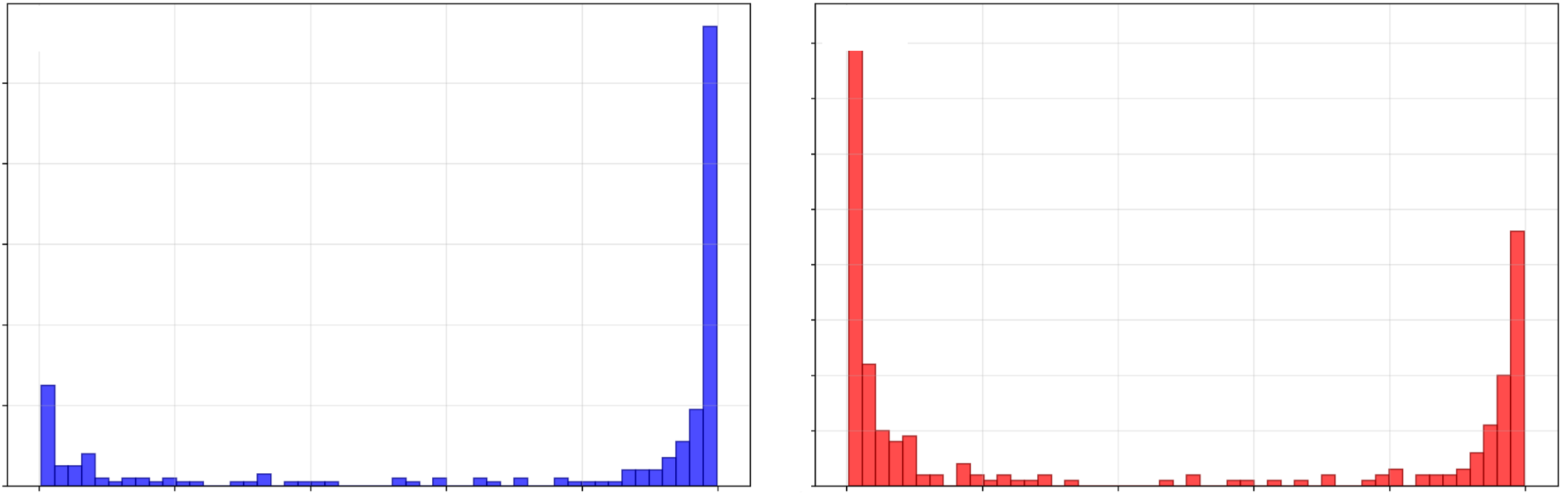}
        \\[-0.5ex] 
        {\footnotesize \makebox[0.85\textwidth][s]{0 \hspace{5em} 1 \hspace{0.1em} 0 \hspace{5em} 1}}
        
        \includegraphics[width=0.85\textwidth, trim={0pt 0 0 0}, clip]{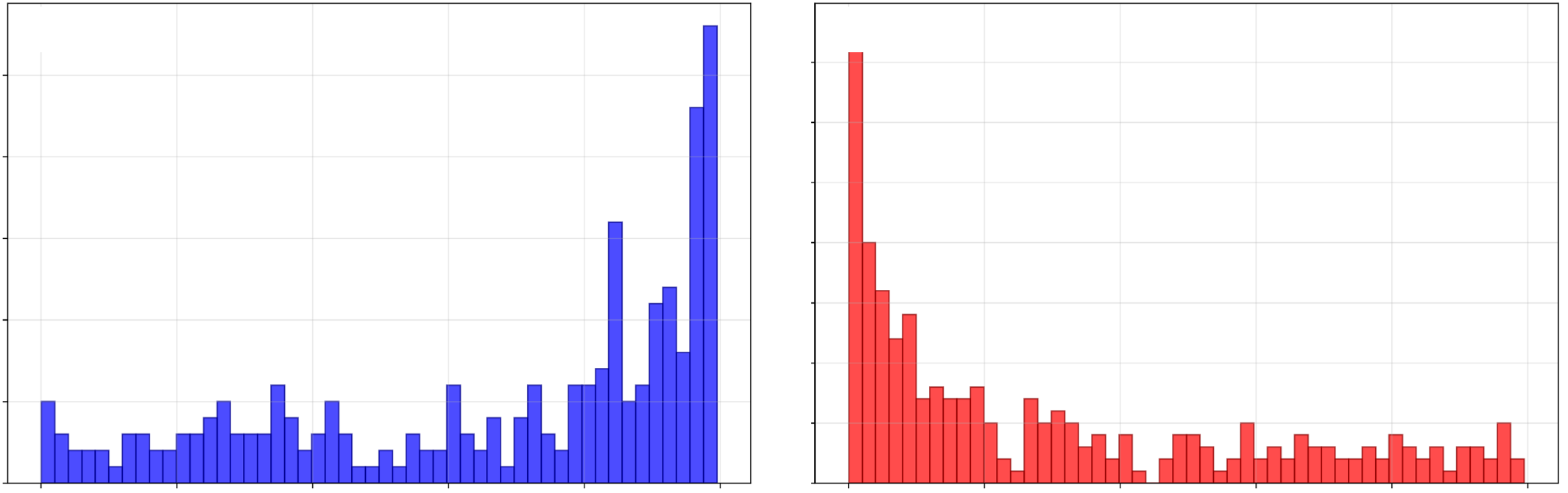}
        \\[-0.5ex]
        {\footnotesize \makebox[0.85\textwidth][s]{0 \hspace{5em} 1 \hspace{0.1em} 0 \hspace{5em} 1}}
        
        \caption{Open-set score distribution histograms}
        \label{fig:hist_group}
    \end{subfigure}

    \caption{Qualitative comparison between the SAFSAR Softmax baseline and FR-Disc on the SSv2 test set. 
    (a) t-SNE of video features $\phi(x_i)$ for the baseline (left) and FR-Disc (right). 
    (b) Score histograms for known (\textcolor{blue}{blue}) and unknown (\textcolor{red}{red}) queries, comparing the baseline (top) and FR-Disc (bottom). 
    Our method yields tighter feature clustering and assigns lower confidence to unknown samples.}
    \label{fig:full_qualitative}
\end{figure*}

\noindent \textbf{Garbage Class: Difficulty of learning a prototype for unknowns.}
The Garbage Class (GC) technique exhibits high sensitivity to both dataset characteristics and backbone architecture. For SAFSAR, we observe a clear dichotomy: while GC provides gains on complex spatio-temporal datasets like SSv2 and Diving48, it consistently degrades performance on datasets with strong spatial biases, such as HMDB51 and UCF101. On our experiments we noticed that on these smaller datasets, the VideoMAE backbone rapidly reaches $100\%$ training accuracy, signaling a collapse into overfitting (see Supplementary Materials for the corresponding convergence plots). We hypothesize that in these regimes, the ``garbage'' category fails to generalize because it only learns to model the ``none-of-the-above'' samples for that specific training set, memorizing static scene cues rather than a general concept of ``unknown'' actions. 

Unlike SAFSAR, STRM suffers a catastrophic drop in Few-Shot Accuracy ($10\%$-$20\%$) on NTURGBD and SSv2 when the GC is added, alongside a general decline in all open-set metrics. This suggests that the explicit modeling of an auxiliary class interferes with STRM’s specific spatio-temporal alignment mechanism by distorting the prototype space during the meta-learning phase.

\noindent \textbf{FR-Disc: Effectiveness of query-prototype feature residuals}.
Among the considered techniques, FR-Disc consistently outperforms on all datasets the Softmax Baseline and the other techniques in both 1- and 5-Shot, closed-set and open-set metrics.
On SAFSAR (Table~\ref{table:safsar}), FR-Disc improves few-shot accuracy in the 5-Shot task by 3.80–5.49\% (0.96\% for UCF101). Regarding open-set performance, the OSCR metric increases by 3.48–5.89\% in the 5-Shot scenario and by 1.40–6.56\% in the 1-Shot scenario, with UCF101 showing consistent gains of approximately 0.70–0.92\%.
When applied to STRM, as shown in Table~\ref{table:strm}, this is the only open-set technique that consistently improves the OSCR metric from the Softmax Baseline in all datasets, with a minor cost of respectively $1.86\%$ and $0.42\%$ of FS ACC on NTURGBD and HMDB51, improving the FS ACC for the other datasets.
This performance gain may be attributed to the fine-grained discrimination capabilities acquired by the Discriminator. There are achieved by the coupling of (i) the known/unknown classification training protocol, that asks to distinguish a query from the most similar known class, and (ii) the residual-based discrimination which, by emphasizing relative variations in the video pair, helps to catch fine distinctions in the feature representation.

\textbf{Improved feature representations.} Figure~\ref{fig:full_qualitative} illustrates how the SAFSAR model benefits from the proposed FR-Disc technique through two qualitative visualizations.
First, Figure~\ref{fig:tsne_group} compares t-SNE~\cite{maaten2008visualizing} projections of the video features $\phi(x_i)$, where each color represents a distinct class.
To ensure a fair comparison, both the Softmax baseline and FR-Disc are evaluated using the identical set of 1000 SSv2 test queries. 
Our method produces visibly more compact clusters with significantly improved inter-class separability.

Second, Figure~\ref{fig:hist_group} displays the open-set score distributions for the same 1000 known (Blue) and 1000 unknown (Red) queries across both configurations. While the Softmax baseline exhibits a high degree of over-confidence—manifesting as a large "acceptance" spike at 1.0 for unknown samples—our FR-Disc method effectively suppresses this peak. 
The resulting scores are better distributed, allowing for a clearer distinction between known classes and outliers.

\section{Conclusion}
This work addresses a critical gap in evaluating few-shot action recognition (FS-AR) methods under open-world conditions, where models must recognize known classes while rejecting unknowns. We introduce the first standardized benchmark for few-shot open-set action recognition (FSOS-AR), adapting five established datasets to represent open-set tasks. By considering a meta-learning prototype-based approach, we demonstrate that closed-set models such as SAFSAR and STRM can be adapted to open-set tasks using techniques from image-based recognition, without sacrificing closed-set accuracy. Notably, for this category of models, we observe that strong closed-set performance often correlates with open-set robustness. Our evaluation identifies FR-Disc as the most reliable explicit OS technique, consistently outperforming alternatives across datasets. Conversely, we observe unstable behavior in GC, which we attribute to the difficulty of learning a representative prototype for unknowns. The implicit EOS technique proved a good trade-off between simplicity and effectiveness. Overall, our analysis aims to establish FSOS-AR as an essential research direction to bridge the gap between controlled experiments and real-world action recognition systems.

\subsubsection{Acknowledgments} The paper was supported by the Italian National Institute
for Insurance against Accidents at Work (INAIL) ErgoCub CORE Project.

\bibliographystyle{splncs04}
\bibliography{mybibliography}

\end{document}